\documentclass[letterpaper, 10 pt, conference]{ieeeconf}  

\IEEEoverridecommandlockouts                              

\usepackage[utf8]{inputenc}
\usepackage{hyperref}
\usepackage{url}
\usepackage{amsfonts}
\usepackage{amsmath}
\usepackage{algorithm}
\usepackage{algorithmicx}
\usepackage{xcolor}
\usepackage{comment}
\usepackage{graphicx}
\usepackage{physics}
\usepackage{multirow}
\usepackage{tikz}
\usepackage{ifthen}
\usepackage{adjustbox}
\usepackage{lipsum}
\usepackage{tabularx}
\usepackage{calc}

\usepackage{svg}
\usepackage{cite}

\title{\LARGE \bf
DeFIX: Detecting and Fixing Failure Scenarios with Reinforcement Learning in Imitation Learning Based Autonomous Driving
}

\author{Resul Dagdanov$^{\ast}$, Feyza Eksen$^{\ast}$, Halil Durmus, Ferhat Yurdakul and Nazim Kemal Ure%
\thanks{$^\ast$ These authors contributed equally to this work}
\thanks{R. Dagdanov is with ITU Artificial Intelligence and Data Science Research Center and Department of Aeronautical Engineering,
        Istanbul Technical University, Turkey
        {\tt\small dagdanov21 at itu.edu.tr}}
\thanks{F. Eksen is with ITU Artificial Intelligence and Data Science Research Center and Department of Computer Engineering,
        Istanbul Technical University, Turkey
        {\tt\small eksen20 at itu.edu.tr}}
\thanks{H. Durmus is with Eatron Technologies and Department of Electronics and Communication Engineering,
        Istanbul Technical University, Turkey
        {\tt\small durmush at itu.edu.tr}}
\thanks{F. Yurdakul is with ITU Artificial Intelligence and Data Science Research Center and
        Department of Aeronautical Engineering, Istanbul Technical University, Turkey
        {\tt\small yurdakul17 at itu.edu.tr}}
\thanks{N.K. Ure is with ITU Artificial Intelligence and Data Science Research Center and Department
        of Aeronautical Engineering, Istanbul Technical University, Turkey
        {\tt\small ure at itu.edu.tr}}%
}

\begin{document}

\maketitle
\thispagestyle{empty}
\pagestyle{empty}

\begin{abstract}
Safely navigating through an urban environment without violating any traffic rules is a crucial performance target for reliable autonomous driving. In this paper, we present a  Reinforcement Learning (RL) based methodology to DEtect and FIX (DeFIX) failures of an Imitation Learning (IL) agent by extracting infraction spots and re-constructing mini-scenarios on these infraction areas to train an RL agent for fixing the shortcomings of the IL approach. DeFIX is a continuous learning framework, where extraction of failure scenarios and training of RL agents are executed in an infinite loop. After each new policy is trained and added to the library of policies, a policy classifier method effectively decides on which policy to activate at each step during the evaluation. It is demonstrated that even with only one RL agent trained on failure scenario of an IL agent, DeFIX method is either competitive or does outperform state-of-the-art IL and RL based autonomous urban driving benchmarks. We trained and validated our approach on the most challenging map (Town05) of CARLA simulator which involves complex, realistic, and adversarial driving scenarios. The source code is publicly available at \url{https://github.com/data-and-decision-lab/DeFIX}
\end{abstract}
\begin{keywords}
Imitation Learning, Reinforcement Learning, Autonomous Driving
\end{keywords}

\section{INTRODUCTION}
Existing approaches to autonomous driving \cite{alizadeh2019automated} have been successful in highly structured environments, such as highways. The next frontier in autonomous driving technology is navigating through the populated urban environment without violating any traffic laws. Driving safely through a highly complex urban environment is a challenging problem. The main issue is a development of an algorithm or a method that can overcome all confronted decision uncertainties while driving. CARLA Challenge \cite{carla} environment is an Unreal Engine based state-of-the-art software simulation that features adversarial urban driving scenarios, e.g. unexpected pedestrians emerge to cross the road from random locations, vehicles run in red lights, occurrence of static barrier or obstacles on the road, decision making in the un-signalized intersections, other vehicles get stuck on the road. Applying Imitation Learning (IL) by mimicking a rule-based policy is insufficient to tackle realistic adversarial scenarios, since these scenarios require complex decision making which would necessitate a lot of hard-coding. As IL agent's performance can be at most as good as the demonstrator policy, use of Reinforcement Learning (RL) could result in higher performance \cite{toromanoff2020end}, because RL agents learn from trial and error by using exploratory behavior and has the potential to surpass the performance of rule-based policy by discovering new patterns between observations and actions. While existing methods assume that a rule-based policy is a perfect demonstrator \cite{chen2020learning, chekroun2021gri}, IL agents may not always carry out safe behaviors on complex decision making scenarios.

\begin{figure*}[t!]
    \centering
    \includegraphics[width=\textwidth]{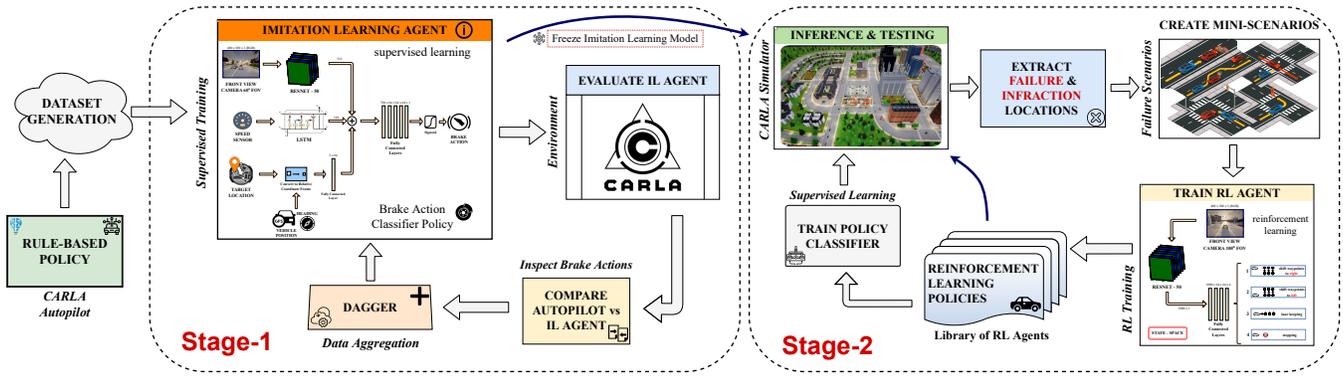}
    \caption{{\bfseries{DeFIX Method Overview}}. {\bfseries{Stage-1}}: An initial dataset is generated by driving a rule-based CARLA autopilot to train an IL agent. This agent is continuously improved with DAgger \cite{dagger} approach. {\bfseries{Stage-2}}: Mini-scenarios are constructed out of infraction and failure locations by evaluating DAgger cantilevered IL policy. At each continual loop of this stage, one different RL agent is trained on one of these mini-scenarios and added to the library of RL policies. In order to activate an IL agent or one of RL agents during evaluation, the policy classifier network is trained with supervised learning.}
    \label{fig:introduction}
\end{figure*}

In this work, we demonstrate that a rule-based policy could also fail and end up obtaining infractions in complex, adversarial scenarios of a realistic urban driving environment. As illustrated in Fig. \ref{fig:introduction}, we present a novel continual learning method, where the failure locations are identified by evaluating trained policies so far and different RL agents are trained on these mini-scenarios where failure cases are re-constructed. A policy classifier network is designed to learn effectively switching between trained RL policies and the baseline IL policy. After training new RL agent on a mini-scenario and adding this trained model to the library of policies, the policy classifier network is trained again from scratch in a supervised manner to avoid the usage of privileged ground truth information from the simulator.

{\bfseries{Contributions:}}
Our main contribution is the DEtect and FIX (DeFIX) framework, which continuously improves the performance of autonomous driving policies by identifying failure cases and replacing them with RL-based agents.

\begin{enumerate}
\item We demonstrate that training an RL agent independently on the extracted mini-scenarios where previously trained agents fail or are unable to handle the given adversarial tasks overcomes given challenges by effectively associating the IL and RL agents together.

\item We propose a supervised policy classifier module that effectively decides on which trained policy to activate during evaluation. This switching mechanism successfully boosts IL agent performance on the given tasks.

\item Even with only one RL agent trained on a failure scenario of an IL agent, we achieve $15\%$ better route completion average score and $3\%$ closer driving score on average than closest benchmark counterpart \cite{chitta2021neat} on the most challenging map (Town05) of CARLA simulator involving complex adversarial urban scenarios.

\end{enumerate}

\section{BACKGROUND}
In Imitation Learning (IL), an agent is trained in a supervised manner to learn the mapping between observations and actions through rule-based demonstrations. More specifically, IL is a way of cloning the expert behavior by treating the actions of a privileged policy as the ground truth labels for each observation. In the domain of autonomous driving, particularly Learning by Cheating (LBC) \cite{chen2020learning} applied an IL method in two steps, training privileged agents using ground truth information from the simulator and training a policy agent by cloning the behavior of a privileged agent. For effective utilization of expert demonstrator, as presented in \cite{bicer2019sample}, the usage of Data Aggregation (DAgger) \cite{dagger} technique in end-to-end autonomous driving tasks yields better performance boosts with fewer data required.

In autonomous driving, Reinforcement Learning (RL) methods are implemented in tactical scenarios such as performing automated overtaking and lane changing by adding safety reward feedback \cite{yavas2020new}, making low-level action decisions \cite{toromanoff2020end} by discretizing throttle and steering commands. The goal of the RL agent is to maximize cumulative reward (a numerical value assigned from an action outcome) by learning from experiences obtained from interacting with the environment. Although RL works well for specific scenarios/tasks, optimizing a policy for generalized urban driving scenarios is impractical, as the required data/experience would be enormous. To effectively separate complex tasks, recent research on safety and efficiency of RL method \cite{curriculum2021rl} is investigated with curriculum learning. Unlike discovering sequential structure between road and weather conditions, applying curriculum to adversarial urban driving tasks is infeasible as deciding on curriculum relationship between these complex and realistic tasks is an open research topic. In a recent work \cite{chekroun2021gri}, the combination of IL and RL increased data efficiency and applied knowledge from demonstrations with exploration. However, they assume that the rule-based policy represents a perfect behavior and that those actions should result in a constant high RL reward.

\section{METHODOLOGY}
In this section, the elements in the proposed DeFIX method, shown in Fig. \ref{fig:introduction}, are explained in detail. 
\subsection{Imitation Learning Architecture}
\begin{figure*}[t!]
    \centering
    \includegraphics[width=\textwidth]{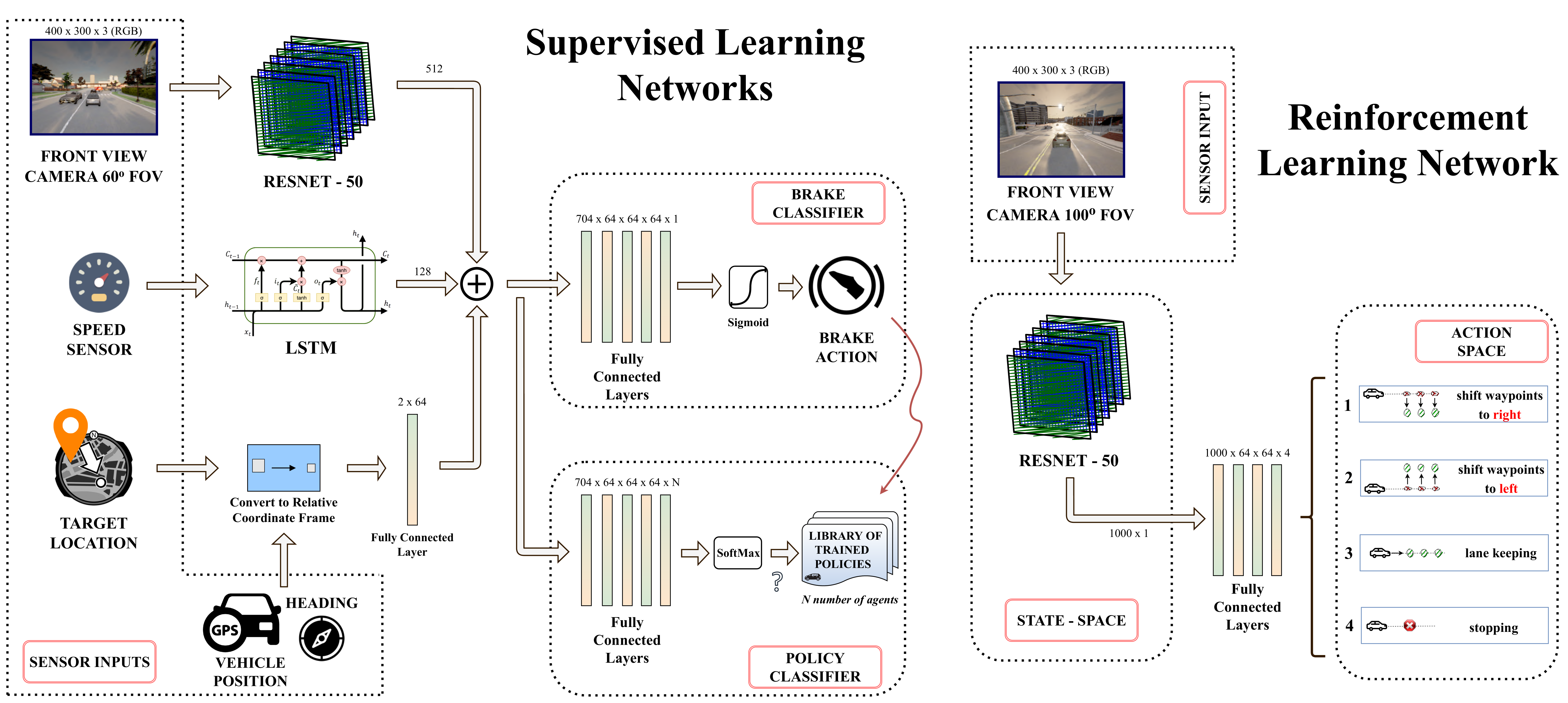}
    \caption{{\bfseries{Network Architectures}}. Pre-trained ResNet-50 \cite{resnet50} model is used as an image backbone to all proposed networks. The last layer of a ResNet-50 is reinitialized and trained from scratch during the training of brake and policy classifiers. Speed, orientation, location, and front camera data are obtained from sensors while sequential target locations are given priorly. Target locations are converted to the relative local coordinate frame by using vehicle position and orientation obtained from IMU and GPS sensors, respectively. In the supervised learning network, a 704-dimensional feature vector represents the fused information of sensor inputs, from there it is processed through Fully Connected layers. IL agent only decides when to apply a brake action while policy classifier decides which trained agent to activate during evaluation. ResNet-50 model is completely frozen during RL trainings. The state-space for the RL agent is a 1000-dimensional vector of ResNet backbone output. RL agents output high-level action commands (lane keeping, right and left lane changing, stopping). Low-level steering and throttle actions are determined with lateral and longitudinal PID controllers, respectively.}
    \label{fig:methodology}
\end{figure*}

A detailed schematics and network flow of our IL model are illustrated in Fig. \ref{fig:methodology}. Initially, a rule-based autopilot is used to collect dataset $D = \{(s^i, a^i)\}_{i=1}^N$ of size $N$. The state space $s \in S$ of the vehicle consists of the front view camera output, target positions, and speedometer outputs where $S$ is the observation set of all possible state spaces. Inertial Measurement Unit (IMU) and Global Positioning System (GPS) outputs are used to determine the orientation and distance to the target waypoints. An objective is to minimize binary cross-entropy loss $\mathcal{L}_{bce}$ of the brake action classifier policy $\pi$ by imitating rule-based policy ${\pi}^*$ which is built upon privileged ground truth information of the simulation.
\begin{equation}
\label{eqn:bce_loss}
    \begin{split}
        \mathcal{L}_{bce}(\pi, {\pi}^*) = & -\frac{1}{M} \sum_{i=1}^M \pi^*(s_i) \cdot log(\pi(s_i)) \\
                                        & + (1 - \pi^*(s_i)) \cdot log(1 - \pi(s_i))
    \end{split}
\end{equation}
where, $M$ is the number of observation states in a mini-batch.

\subsubsection{Network Architecture}
As shown in Fig. \ref{fig:methodology}, image features coming from the ResNet-50 backbone are concatenated with positional embeddings and speed features. The image input resolution of the IL model is 400 x 300 RGB in pixels and the output size of the backbone network is 512. GPS coordinates of the target waypoints are converted into a local frame and feed into the Fully Connected (FC) layer with activation of Rectified Linear Units (ReLU) outputting positional embeddings of size 64. Ego vehicle velocity features are extracted from the Long Short-Term Memory (LSTM) layer of hidden size 128 and sequential input size 120. As illustrated in Fig. \ref{fig:methodology}, a brake command is seen as a $"True-False"$ action, so a Sigmoid activation function is used. $True$ means applying maximum brake action and $False$ means ignoring brake action and fully relying on outputs of the low-level throttle and steer PID controllers.

\subsubsection{Throttle and Steering Controllers}
\label{sssection: pid_controllers}
Both IL and RL agents rely on low-level controllers for throttle and steering actions. We interpret the throttle action as an absence of a brake command and vice-versa, so the action space size of the agent vehicle is limited to 2 actions (throttle-brake and steering). A fixed set of long-distance waypoints $\vb*{\vu{w}} = \{\hat{w}_1, \hat{w}_2, ..., \hat{w}_C\}$ are given prior to evaluation. The distance between given GPS coordinates of target waypoints is in a range between 7.5 meters and 50 meters. We interpolate long-distance given waypoints so that the short-distance between any consecutive target position node is approximately 4 meters. These parameters are common choices and they are fixed for usage in CARLA competition.

In our framework, two different PID controllers, similar to \cite{chen2020learning}, are used in the determination of steering and throttle actions. In this work, the same controllers and fixed parameters are applied to all agents including the rule-based autopilot which has access to privileged information.

A longitudinal PID controller tries to minimize the error between target velocity $v^*$ and agent vehicle's current velocity $v_{agent}$. The calculation of ego vehicle target velocity is given in Eq. (\ref{eqn:target_velocity}). This equation represents the average sequential velocities though each consecutive waypoints given $\Delta t$ time.
\begin{equation}
\label{eqn:target_velocity}
    v^* = \frac{1}{C} \sum_{i=1}^C \frac{||\hat{w}_i - \hat{w}_{i-1}||^2}{\Delta t}
\end{equation}

A lateral PID controller tries to minimize the error between the agent vehicle's current steering angle $\alpha$ and the target steering angle $\alpha^*$. The arc is fitted between consecutive short-distance waypoints to increase accuracy. The calculation of target steering angle is given as $\alpha^* = \arctan \frac{T_y}{T_x}$, where $T$ is the nearest arc position node. During evaluation and RL agent training, nearest waypoint node $\hat{w}_i$ to ego vehicle is assigned to a target position node $T_i$ as $i \to C$ where $C$ is total number of waypoints in one route.

\subsection{DQN Architecture}
We use the RL method, specifically the Deep Q Network (DQN) agent, to compensate the behavior of the trained agents and to learn a safe decision policy in complex scenarios where previous agents get infractions or fail to navigate through given routes safely.

\subsubsection{State-Space}
The observation state of our RL agents is a 1000-dimensional vector, an output of a pre-trained and fully-frozen ResNet-50  \cite{resnet50}. The input of the ResNet-50 is the front camera image of size 400 x 300 RGB with $100$ degrees of field-of-view (FOV). During backpropagation through DQN, only the weights of the FC layers, in Fig. \ref{fig:methodology} DQN part, are optimized.

\subsubsection{Action Selection}
We generalize the action space of RL agents with high-level commands which are braking, lane changing to left, lane keeping, and lane changing to right as illustrated in Fig. \ref{fig:methodology}. Detailed clarification of throttle and steering PID controllers is given in Section \ref{sssection: pid_controllers}. Lane change commands are defined by shifting the given target waypoints $\pm 3.5$ meters with $\pm 90^{\circ}$ relative to the heading angle of an agent vehicle as depicted in RL part of Fig. \ref{fig:methodology}.

\subsubsection{Reward Function}
To enclose infraction metrics of all realistic scenarios, we design detailed and generalized reward function $\mathcal{R}(s_i)$. This reward function is also used in the determination of target labels of the policy classifier as it outputs negative feed-backs from possible traffic infractions. This function is used for all RL agents as it reveals penalties for possible infraction cases and positive rewards for correct actions. The energy weights of $\mathcal{R}(s_i)$ are determined with experiments and emphasises on the importance.
\begin{equation}
\label{eqn:reward_function}
     \begin{split}
        \mathcal{R}(s_i) = & \delta(s_i) + [1 - \xi_{c_s}] \cdot [(1 - \phi(s_i)) \cdot V(s_i) \\
        & + 50 \cdot \phi(s_i) \cdot \beta(s_i) - \phi(s_i) \cdot V(s_i)] \\
        & + \xi_{c_s} \cdot V(s_i) - 100 \cdot \tau(s_i) - 1500 \cdot \zeta(s_i)
    \end{split}
\end{equation}
where,
\begin{itemize}
    \item $s_i$: \quad privileged observation space at a given step $i$
    \item $V(s_i)$: \quad speed of the agent vehicle in $m/s$
    \item $\xi_{c_s}$: \quad $1$ - if $\hat{c}(s_i, t_i)=2$; $0$ - otherwise
    \item $\beta(s_i)$: \quad brake command of an agent vehicle [$0$ or $1$]
    \item $\phi(s_i)$: \quad $1$ - if an affecting traffic light is red or yellow, existence of vehicles and pedestrians are in the field of influence of a rule-based agent; $0$ - otherwise
    \item $\delta(s_i)$: \quad Euclidean distance between agent's location and latest passed waypoint location in $meters$
    \item $\tau(s_i)$: \quad $1$ - if agent velocity is $0.0$ $m/s$ for $60$ consecutive seconds; $0$ - otherwise
    \item $\zeta(s_i)$: \quad boolean function revealing any collisions
\end{itemize}

\subsection{Policy Classifier}
\label{ssection: policy_classifier}
The purpose of policy classifier is to switch policies dynamically between a library of trained RL agents and IL agent during evaluation. As given in Fig. \ref{fig:introduction} Stage-2, the policy classifier is trained from scratch in a continuous loop after each new trained DQN agent addition to the library of RL policies. As shown in Fig. \ref{fig:methodology}, an image backbone network (ResNet-50) and input observation states are the same as in IL agent, however, IL and policy classifier networks are trained separately. A policy classifier network's loss is calculated with multi-class cross-entropy loss function $\mathcal{L}_{mce}(\pi, {\pi}^*) = -\frac{1}{M} \sum_{i=1}^M \frac{1}{N} \sum_{j=1}^N \pi^*(s_i)_j \cdot log(\pi(s_i)_j)$ where $M$ is the number of observation states in the mini-batch and $N$ is the number of distinct policy classes. As illustrated in Fig. \ref{fig:introduction}, we completely rely on classifier outputs and do not use any privileged information during evaluation.

The target ground truth labels for policy classifier are defined as given in Eq. (\ref{eqn:policy_classes}) where the positive return of Eq. (\ref{eqn:reward_function}) represents correct action performance of an IL agent without any traffic rule infractions. By applying IL policy and getting negative reward for a given state $s_i$, shows that an infraction is occurring and the switch to another policy should be held.

\begin{equation}
\label{eqn:policy_classes}
    label(s_i, \pi) =
    \begin{cases}
        \text{{\textit{IL Agent}}}, &\text{if} \quad \mathcal{R}(s_i) > 0.0 \\
        \hat{c}(s_i, \pi), &\text{if} \quad \mathcal{R}(s_i) \leq 0.0 \\
    \end{cases}
\end{equation}
where $\hat{c}(s_i)$ represents an auto-labeled failure scenario as a privileged reward function $\mathcal{R}(s_i)$ outputs negative value at an agent's current state $s_i$ under evaluated policy $\pi$.

\section{EXPERIMENTS}
\subsection{Simulation Setup}
\subsubsection{Preprocessing}
The processed raw images are normalized to the range [0, 1] by dividing all pixels by $255$. The camera module is placed at $2.3$ meters of height from the ground and $1.3$ meters in front of the ego vehicle's center. These parameters are chosen as they are a common choice in CARLA Challenge \cite{chitta2021neat, toromanoff2020end, prakash2021multi} environment.

\subsubsection{CARLA Autopilot}
A rule-based autopilot policy acquires the privileged information from the simulation. The autopilot has lateral and longitudinal PID controllers that generate low-level steering and throttle commands from the given navigation waypoints. Similar controllers were used in \cite{chitta2021neat, toromanoff2020end, chen2020learning, prakash2021multi} and detailed description of these controllers are reviewed in Section \ref{sssection: pid_controllers}. On top of these PID controllers, brake command is independently activated/deactivated based on the ground truth information. This information includes current distance and orientation difference between the closest vehicle, object, pedestrian, and affecting traffic light to the autopilot vehicle. The same autopilot is used in \cite{prakash2021multi} for rule-based demonstrator data collections.

\subsubsection{Dataset Generation}
CARLA Challenge provides several publicly available default towns. We trained and tested our agents on the most challenging map \cite{toromanoff2020end} Town05, that includes multi-lane highways and US-style traffic lights. Town05 is the biggest distinct out of the publicly available maps with 9 and 32 total number of different route trajectories in Town05-Long and Town05-Short. With the purpose of generalization, weather conditions are changed every 20 time-steps randomly across various weather states (clear, wet, cloudy, soft-rainy, hard-rainy) and daytimes (noon, sunset) during data collections. Image, control output, GPS coordinate of the nearest target waypoint, speed sequence, and a reward function Eq. (\ref{eqn:reward_function}) values are stored every 10 simulation steps. A simulation frequency is fixed to 20 Hz.

\subsubsection{Scenarios}
\label{sssection: scenarios}
CARLA Challenge includes adversarial driving scenarios which are emerging pedestrians to cross the road, vehicles running in red lights, uncontrolled turns at intersections, fixed obstacles on the waypoints, stuck vehicles on the driveway etc. We tested our framework in these scenarios and evaluated the performance of our trained policies alongside a policy classifier network.

\subsection{Training}
\subsubsection{Imitation Learning Agent Training}
Initial IL training is done with approximately $22K$ CARLA rule-based autopilot demonstration samples $D_0$. By evaluating this trained IL agent with the DAgger method \cite{dagger} until completing all 41 routes of Town05, approximately $2K$ samples are collected and aggregated to initial $D_0$ to generate a dataset $D_1$. DAgger samples are collected when IL agent's brake command policy $\pi(s)$ shifts from an autopilot's brake action policy $\pi^*(s)$ for an observed state $s$. After training the first DAgger agent with on $D_1$, the second iteration of samples is collected in the same manner and the performance completions of each IL policy is shown on Table \ref{table:dagger_results}.

\subsubsection{Reinforcement Learning Agent Training}
In the switch to the inference and testing section illustrated in Stage-2 of Fig. \ref{fig:introduction}, the IL agent, which is trained on $D_2$, is evaluated on Town05 routes. Failure and infraction locations of IL agent's performance are extracted. Initially, one mini-scenario is constructed on the stuck vehicle scenario of Town05-Long from IL agent testings. This stuck vehicle scenario is chosen randomly out of all similar failure cases of IL agent on Town05-Long and Town05-Short routes. Blue framed image on Fig. \ref{fig:stuck_scenario_illustrations} shows the training scenario while orange framed images show some of other stuck vehicle failures.

\definecolor{bluebell}{rgb}{0, 0, 0.4}
\definecolor{orangebell}{rgb}{1, 0.5, 0}
\begin{figure}[H]
    \centering
    \includegraphics[width=0.44512\textwidth]{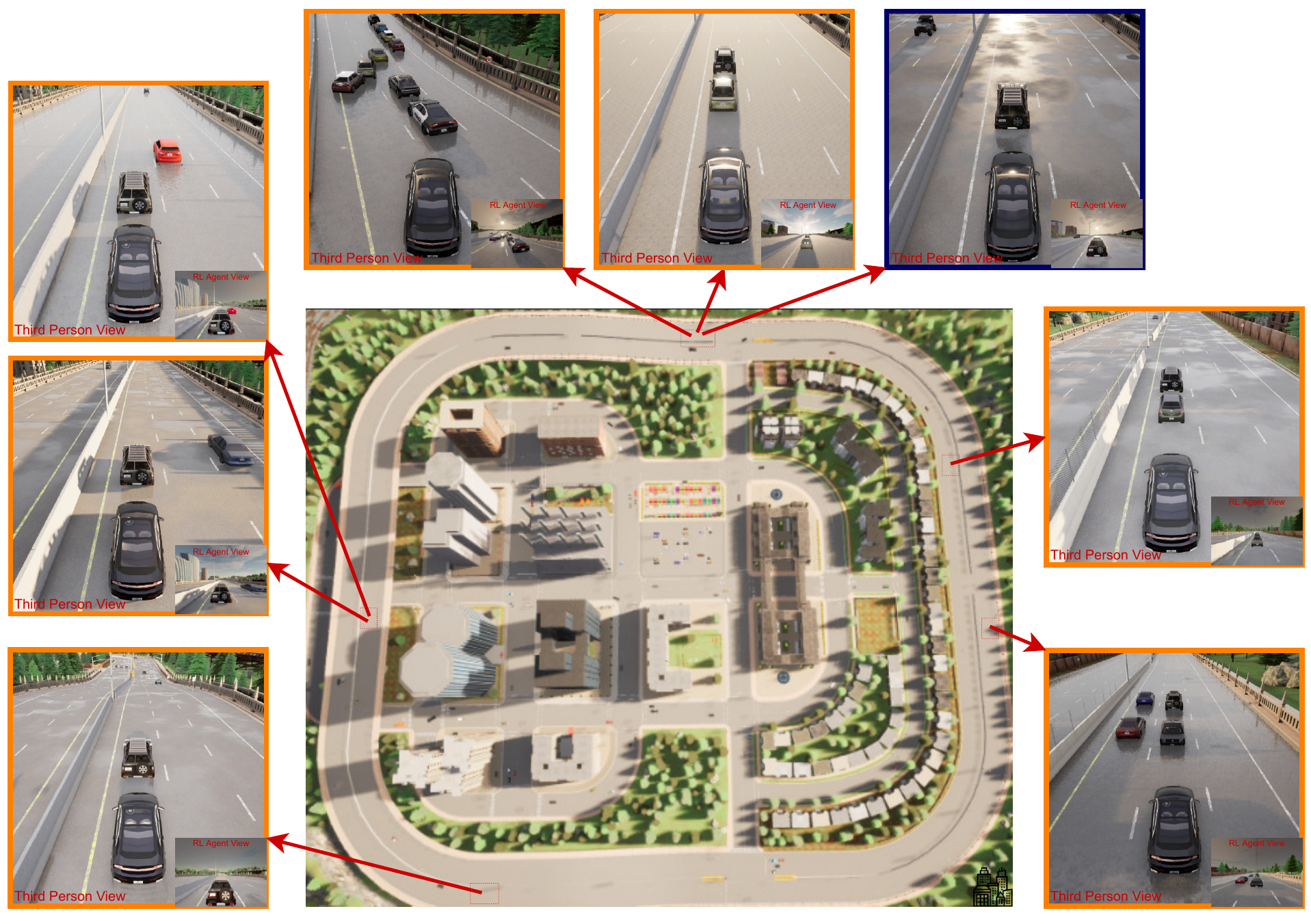}
    \caption{Illustration of stuck vehicle scenarios in Town05. {\bfseries{{\color{bluebell} Blue}}} boxed scene is a training scenario of the RL policy that aims to optimize an overtaking decision. {\bfseries{{\color{orangebell} Orange}}} boxed scenes are the examples of evaluation scenarios.}
    \label{fig:stuck_scenario_illustrations}
\end{figure}

A detailed reward function, given in Eq. (\ref{eqn:reward_function}), is used to train RL agents. The training episodes are terminated when the maximum step number of an episode (250 steps) is reached or a collision of any kind occur ($\zeta(s_i)=1$) as depicted in Eq. (\ref{eqn:reward_function}). To generalize, weather conditions are dynamically changed at every 20 steps and the RL agent is initialized at a distance of range [3 meters - 10 meters] relative to the exact failure location during training.

\subsubsection{Policy Classifier Training}
At each state $s_i$ under the evaluated policy $\pi$, a failure scenario label $\hat{c}(s_i, \pi)$ is determined with a negative sign of a generalized reward function $\mathcal{R}(s_i)$ as depicted in Eq. (\ref{eqn:policy_classes}). This reward function, that is given in Eq. (\ref{eqn:reward_function}), uses privileged information from the simulator regarding the existence and state of affecting traffic lights, collision detection label, etc. During data collection for both policy and brake classifier networks, privileged ground truth auto-labeled information is used. However, during inference and evaluation, our framework does not require any privileged information as action prediction and policy classifications rely on outputs of trained networks which fully rely on sensor inputs.

The policy classifier network is trained in a supervised learning manner as described on Stage-2 of Fig. \ref{fig:introduction}. The ground truth training data samples are collected during the evaluation of trained policies. Initially, only IL agent policy is evaluated and data samples are collected according to Eq. (\ref{eqn:policy_classes}). After training first RL agent on a failure scenario, an initial policy classifier model is trained with 2 classes.

\section{RESULTS \& DISCUSSION}
\subsection{Evaluation Metrics}
The overall performance of DeFIX is analyzed with official metrics of the CARLA Leaderboard. Route Completion (RC) is given in an amount of percentages distance completed on each route of Town05. If an agent deviates from the given route or gets blocked by stuck vehicles, RC will be penalized in completion percentages. Infraction Score (IS) is calculated as an aggregated penalty for every traffic rule infraction, red light violation, lane infraction, or collision with vehicles and pedestrians. Agent receives a maximum of 1.0 IS when no infraction is done during evaluation. Driving Score (DS) is computed as a product of IS and RC throughout all routes in Town05-Long and Town05-Short.

\subsection{Benchmark Results}
Table \ref{table:dagger_results} shows DS, IS, and RC performance scores of initial IL and DAgger agents with rule-based demonstrator policy on all routes of Town05. By only imitating a rule-based agent's boolean brake action, we received close RC scores to a privileged autopilot policy. As RC scores of Table \ref{table:dagger_results} illustrate, even after several DAgger iterations, an IL agent could be as good as a demonstrator policy at best.

\begin{table}[ht]
\caption{Imitation Learning Agent Performance on Town05}
\begin{adjustbox}{width=\columnwidth}
    \begin{tabular}{ c | c | p{1.22cm} | p{1.22cm} | p{1.22cm}}
        \cline{3-5}
            \multicolumn{1}{l}{} & \multicolumn{1}{l}{} & \multicolumn{3}{p{4.2cm}}{\qquad \qquad Performance Metrics}\\
        \hline \hline
            {\bfseries{Town05-Short}} & Trained Dataset & \quad {\bfseries{RC}} $\Uparrow$ & \quad {\bfseries{IS}} $\Uparrow$ & \quad {\bfseries{DS}} $\Uparrow$ \\
        \hline
            Imitation & $D_0\rightarrow22K$ & 89.8$\pm$2.27 & 0.74$\pm$0.01 & 64.95$\pm$1.3 \\
            Learning & $D_1\rightarrow24K$ & 90.0$\pm$2.57 & 0.68$\pm$0.01 & 62.91$\pm$3.4 \\
            Agent & $D_2\rightarrow26K$ & {\bfseries{90.1}}$\pm$1.12 & {\bfseries{0.76}}$\pm$0.02 & {\bfseries{68.47}}$\pm$2.9 \\
        \hline
            {\textit{Autopilot Agent}} & $N/A$ & \quad 90.94 & \quad 0.91 & \quad 82.75 \\
        \hline \hline \hline
            {\bfseries{Town05-Long}} & Trained Dataset & \quad {\bfseries{RC}} $\Uparrow$ & \quad {\bfseries{IS}} $\Uparrow$ & \quad {\bfseries{DS}} $\Uparrow$ \\
        \hline
            Imitation & $D_0\rightarrow22K$ & 75.1$\pm$0.33 & 0.39$\pm$0.09 & 25.59$\pm$4.8 \\
            Learning & $D_1\rightarrow24K$ & 75.4$\pm$0.03 & 0.41$\pm$0.03 & 30.91$\pm$2.2 \\
            Agent & $D_2\rightarrow26K$ & {\bfseries{75.4}}$\pm$0.06 & {\bfseries{0.44}}$\pm$0.03 & {\bfseries{33.17}}$\pm$2.2 \\
        \hline
           {\textit{Autopilot Agent}} & $N/A$ & \quad 75.41 & \quad 0.693 & \quad 48.60 \\
        \hline
    \end{tabular}
\end{adjustbox}
\label{table:dagger_results}
\end{table}

As seen in Table \ref{table:benchmark}, IL agent's RC and DS performances are boosted even with only one RL agent that is trained on the stuck vehicle infraction that is represented in Fig. \ref{fig:stuck_scenario_illustrations}. Results of World on Rails \cite{chen2021learning} and NEAT \cite{chitta2021neat} agents are obtained on the same scenarios over 9 runs while results of other benchmark agents are taken from \cite{prakash2021multi}. We point out the evaluation performance of a single RL agent that is trained on a stuck vehicle scenario alone on all routes and scenarios.

\begin{figure}[ht]
    \centering
    \includegraphics[width=0.485\textwidth]{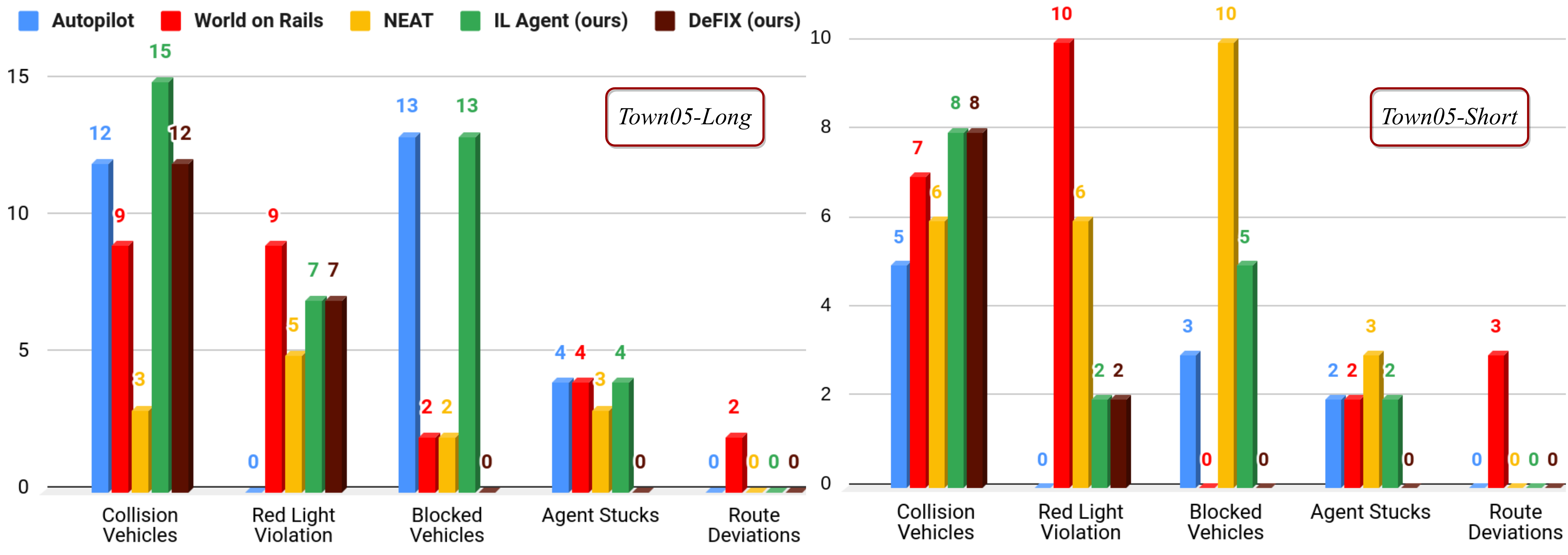}
    \caption{{\bfseries{Infractions}}. Total number of infractions of listed models are compared with DeFIX (ours) in the Town05 settings where {\textit{Town05-Long}} has $9$ routes and {\textit{Town05-Short}} has $32$ routes. (lower is better)}
    \label{fig:bar_graph_town05}
\end{figure}

\begin{table}[ht]
\caption{Benchmark Comparison of Driving Performance Scores}
\centering
\begin{tabular}{ l | c | c || c | c }
    \hline
        \multicolumn{1}{c|}{\bfseries{Method}} & \multicolumn{2}{c||}{\bfseries{Town05-Short}} & \multicolumn{2}{c}{\bfseries{Town05-Long}}\\
    \hline \hline
        & \quad {\bfseries{RC}} $\Uparrow$ & \quad {\bfseries{DS}} $\Uparrow$ & \quad {\bfseries{RC}} $\Uparrow$ & \quad {\bfseries{DS}} $\Uparrow$\\
    \hline
        LBC \cite{chen2020learning} & 55.01 & 30.97 & 32.09 & 7.05 \\
        Late Fusion \cite{prakash2021multi} & 83.66 & 51.56 & 68.05 & 31.30 \\
        CILRS \cite{codevilla2019exploring} & 13.40 & 7.47 & 7.19 & 3.68 \\
        AIM \cite{prakash2021multi} & 81.07 & 49.00 & 60.66 & 26.50 \\
        TransFuser \cite{prakash2021multi} & 78.41 & 54.52 & 56.36 & 33.15 \\
        NEAT \cite{chitta2021neat} & 69.34 & 58.21 & 88.78 & {\bfseries{57.49}} \\
        Geometric Fusion \cite{prakash2021multi} & 86.91 & 54.32 & 69.17 & 25.30 \\
        World on Rails \cite{chen2021learning} & 52.60 & 38.14 & 60.57 & 32.18 \\
        DeFIX (ours) & {\bfseries{96.34}} & {\bfseries{72.41}} & {\bfseries{89.61}} & 39.42\\
        IL Agent (ours) & 90.10 & 68.47 & 75.40 & 33.17\\
        RL Agent (ours) & 30.14 & 24.65 & 5.17 & 4.15\\
    \hline
    {\textit{Autopilot Agent}} & 90.94 & 82.75 & 75.41 & 48.60\\
    \hline
\end{tabular}
\label{table:benchmark}
\end{table}

\subsection{Discussion}
After comparing our DeFIX model on the state-of-the-art benchmark as shown in Table \ref{table:benchmark}, we conclude that training an RL agent on the failure scenario of the IL model improves general driving performance by outperforming even privileged autopilot in RC score on all routes due to the capability of the trained RL agent that makes overtaking maneuvers in stuck vehicle scenarios. As depicted in Fig. \ref{fig:bar_graph_town05}, our DeFIX methodology successfully reduced the total number of agent stucks and number of blocked vehicles to 0 as RL agent could make overtaking actions while navigating through given routes. Our DeFIX method successfully solved all 6 diverge stuck vehicle scenarios of Town05 while an IL agent and a rule-based autopilot agent failed.

\section{CONCLUSION \& FUTURE WORK}
In this work, we have demonstrated that IL agents are limited with the performance of the demonstrator policy. To further increase the performance while achieving safe autonomous driving, we proposed and tested the DeFIX method. With the combination of DAgger cantilevered IL policy, library of RL agents, and policy classifier, we have achieved better RC and DS than the methods in the literature on Town05 map of CARLA driving simulator. Since DeFIX method employs continual learning, newly encountered scenarios will be solved with robust RL policies and stored in the library to be selected by policy classifier.

The future of this work will focus on extensive tests on different maps and scenarios of the CARLA simulator. The confrontation of new challenges and failure cases will expand the library of RL agents and will broaden the effectiveness of the policy classifier. The evaluation of this method will be carried on CARLA challenge to further investigate the capabilities of our approach. We believe that our DeFIX methodology could spark reliable, robust, and safe autonomous driving technologies.

\section*{ACKNOWLEDGMENT}
This work is supported by Istanbul Technical University BAP Grant NO: MOA-2019-42321. We gratefully thank Eatron Technologies for their technical support. Feyza Eksen thanks the DeepMind scholarship program for their support.

\bibliography{references}

\addtolength{\textheight}{-12cm}  
\end{document}